# Developing an OCR model for Extracting Information from Invoices with Korean Language

Xiem HoangVan[1], Phu TranQuang[1], Minh DinhBao[1], Tien VuHuu[2]
[1] *University of Engineering and Technology, Vietnam National University-Hanoi, Vietnam*
[2] *Posts and Telecommunications Institute of Technology, Hanoi, Vietnam*
*xiemhoang@vnu.edu.vn, trn.quangph27@gmail.com, minhdinh@vnu.edu.vn, tienvh@ptit.edu.vn*

***Abstract*— Invoices are commercial documents that contain various pieces of information, including the purchased items, time, and total money. Making the extraction of important information crucial. The stored information serves different purposes. Korean language is the native language of about 80 million people, playing an important role in not only South and North Korea but also in many other countries such as Vietnam, Philippine where a large number of Korean companies are located. In this context, to automatically extract proper information from the invoices with Korean language, we propose an efficient Optical Character Recognition (OCR) model in which a deep learning model is combined with some image preprocessing techniques. The proposed OCR model is assessed in a rich set of collected invoices showing that 87% F1-score can be achieved with negligible time processing.**

## I. INTRODUCTION

Every day a large number of invoices are sent from suppliers to consumers, which are documents containing and recording information about product transactions. Extracting essential information such as purchased time, consumed products, and total expenditure from invoices plays an important role in commercial activity. Generally, we can perform this task manually by handwriting or typing the necessary information found in the invoice. However, this method is inefficient and time-consuming, especially when dealing with large amounts of transactions.

Additionally, there is an increasing number of foreign companies investing in South East Asia, opening branches, and establishing direct manufacturing plants. South Korea is a notable example [1]. This country is also chosen by many Vietnamese people for studying abroad or working as migrant workers. For these reasons, the extraction and storage of information from invoices during transactions become more important and necessary for people from not only Korea but also relevant countries.

Recognizing the strong rise of Artificial Intelligence (AI), this technology is perfectly suitable and can be applied to the aforementioned practical problem [2]. It can replace humans in quickly identifying and recording the important information that needs to be stored on invoices, ensuring accuracy, cost savings, and labor efficiency. Moreover, AI technology can permanently operate, catering to the specific needs of individuals.

| ㅏ | ㅑ | ㅓ | ㅕ | ㅗ | ㅛ | ㅜ | ㅠ | ㅡ | ㅣ | |
|---|---|---|---|---|---|---|---|---|---|---|
| a | ya | o | yo | ô | yô | u | yu | ư | i | |
| ㅐ | ㅒ | ㅔ | ㅖ | ㅘ | ㅙ | ㅚ | ㅝ | ㅞ | ㅟ | ㅢ |
| e | ye | ê | yê | oa | oe | uê | uơ | uê | uy | ưi |
| ㄱ | ㄴ | ㄷ | ㄹ | ㅁ | ㅂ | ㅅ | ㅇ | ㅈ | ㅊ | ㅋ |
| k/g | n | t/d | r/l | m | b | s | ng | j | ch | kh |
| ㅌ | ㅍ | ㅎ | ㄲ | ㄸ | ㅃ | ㅆ | ㅉ | | | |
| th | p' | h | kk | tt | pp | ss | ch | | | |

Fig. 1. Korean letters

Information from invoices with Korean languages can be extracted with several OCR methods, such as [3, 4]. However, these models were not mainly designed for extracting information from invoices with Korean but for general languages. Therefore, there is still room for further develop the accuracy of OCR model with this type of language if both deep learning and image processing techniques are properly combined. In this context, we proposed a novel OCR model for text recognition in invoices using the Korean language, which is highly applicable in real-world. Specifically, the proposed work includes three main contributions. Firstly, a text detection model with high accuracy is proposed. Secondly, a SVTR architecture is proposed for learning features of Korean characters. Finally, a complete system is developed to extract information from image of Korean invoices.

The rest of this paper is organized as follows. In the Section II, we briefly introduce some related works. Section III describes the proposed OCR model while Section IV presents the performance assessment. Finally, Section V gives some conclusions and future works.

## II. RELATED WORK

### *A. Text Detection*

In recent years, deep learning and computer vision have been used for text detection in various contexts. However, the performance of these models is not good enough when the text in the image is deformed, skewed, or curved. To address these limitations, a model called Character Region Awareness for Text Detection (CRAFT) [5] was proposed with the ability to detect characters in various contexts, performing well on curved, long, and deformed text. Previous models required the generation of ground truth for character-level text processing, which is a costly and time-consuming task. Instead, CRAFT used a convolutional neural network (CNN) to compute region scores for each area, along with scores for the relationships between regions. The region scores are used to reconstruct the spatial map of text regions, while the relationship scores are used to group characters into text regions.

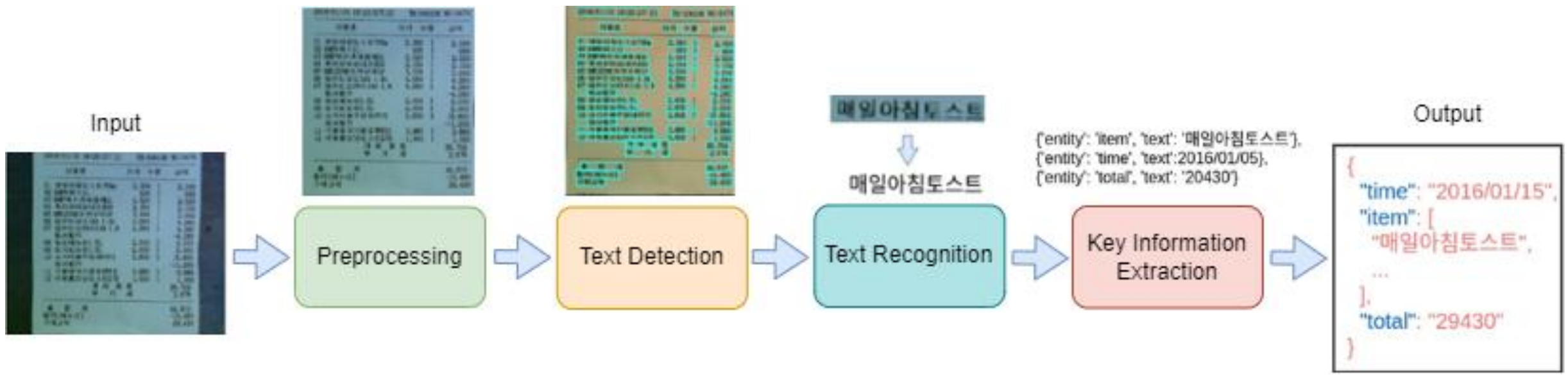

Fig. 2. System design diagram

The neural network used in CRAFT is entirely based on the VGG16 model [5], with the ability to infer text detection at the word level. This requires significant resources for processing and storage, which can be costly and operationally challenging.

### B. *Text Recognization*

Regarding character recognition, this problem is considered a challenging task with great importance. Many research works have been conducted in this area, but the most notable one is the CRNN-CTC model [6]. Compared to conventional character recognition methods that detect the position of each character and recognize them using dynamic programming and search algorithms, these methods cannot exploit the semantic relationships between characters. To address this limitation, the CRNN model adopts the idea of using a convolutional neural network to extract a feature sequence from the image, which is then fed into a recurrent neural network (RNN) or a long short-term memory (LSTM) network to predict each character at each frame.

In CRNN, the CTC function was used as the decoding layer, effectively addresses the limitations or errors that text prediction models often encounter. These errors include word repetition, computational cost limitations, and time-saving benefits. Both the CRNN model and models combining convolutional neural networks with recurrent neural networks generally have the drawbacks of long training times due to sequential computations. Another limitation is their dependency on long-distance information, leading to inferior results for long sentences.

### C. *Key Information Extraction*

In terms of information extraction, apart from using the outputs of the two previous tasks to build features and graphs for training, there are alternative approaches such as Template-based and NLP-based methods.

- *The Template-based* approach applies predefined rules to structured documents with fixed and relatively stable formats [7]. It then utilizes methods such as keyword matching to identify corresponding information. However, this method has limitations as it requires defining individual rules for each specific document structure and heavily relies on the understanding ability of individuals, making it time-consuming.
- *NLP-based*, it belongs to the field of Natural Language Processing (NLP) where textual information is inputted into a text classification model or a Named Entity Recognition (NER) task [8]. The purpose is to classify or identify entities belonging to specific fields of information. Although this method is adaptable to new data, it still depends heavily on the layout of each text, which limits its effectiveness when dealing with data represented in tabular form. Additionally, this method cannot utilize positional features of text cells, despite their significant contribution to determining corresponding information fields.

### D. *Korean language*

Korean is the native language of about 80 million people. Modern Korean is written in the Korean script Hangul in South Korea; Chosŏn'gŭl in North Korea [9]. The script consists of 40 characters with 21 vowels and 19 consonants as depicted in Fig.1.

## III. Proposed Method

The proposed OCR model for extracting the information from invoices with Korean language is illustrated in Fig. 2. First, during preprocessing, the invoice image is background removed and perspective adjusted. This step aims to address issues like image noise, poor quality, and unfavorable camera angles that can impact the accuracy of subsequent processes.

The image then undergoes text detection to locate all the text areas in the image. Text in the detected area then read and identified in the text recognition task. Finally, the key information extraction task for Korean invoices will extract three fields including purchase time, purchased items, and total invoice amount.

### A. *Dataset creation and overal work flow*

Our dataset comprises images of invoices written in Korean. These images are obtained by either capturing physical invoices in Korea or by searching on Google or Naver search engines. Despite the presence of several OCR competitions focused on invoices, particularly in technologically advanced countries like Korea, publicly available datasets specifically tailored for Korean invoices are scarce.

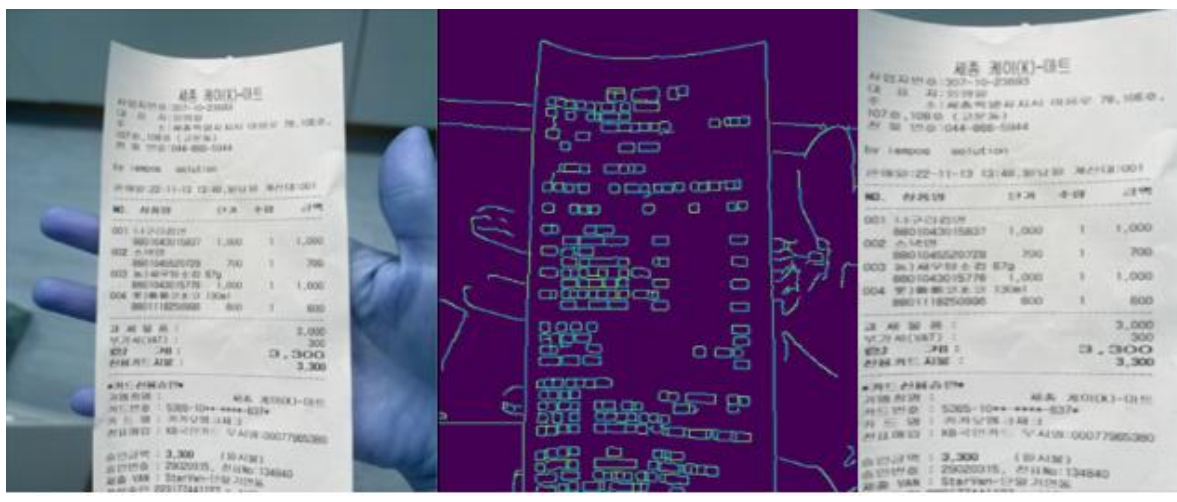
Fig. 3. Preprocessing result

Initially, we collected data from various websites. However, the data faced significant challenges in terms of image quality due to the manual capturing using different devices and varying environments. The quality of the paper used for the invoices also varied, leading to inconsistent and blurry images with unclear text. Additionally, many images were captured from poor angles, which posed challenges for the recognition system. The subpar image quality had a detrimental impact on both the training and prediction processes.

Afterwards, we filtered out low-quality images, opting for a manual filtering approach as our second option. Each image was carefully examined, and those of poor quality were removed. In the end, only 375 images were kept as dataset for our project.

Following the image filtering process, we discovered that many images in the dataset contained unnecessary background information or were captured from unfavorable angles. Therefore, we proceeded to the next step of preprocessing, where we transformed the invoice images to a better perspective and removed any unnecessary background elements (Fig. 3).

After the image processing task is completed, the next step is labeling. We utilized an automatic labeling tool provided by [10]. However, it is important to note that the accuracy of this labeling tool is not entirely satisfactory, necessitating manual intervention to correct any errors. Therefore, we invested additional effort in manually fixing and refining the labels to ensure the accuracy and reliability of the dataset.

### B. *Proposed OCR model*

To extract the intended information in Korean invoices, we perform the following steps.

#### 1. *Preprocessing*

To identify invoices in images, the Canny edge detection algorithm [11] is used. Subsequently, the contours which are curves that connect contiguous points with the same magnitude value is located. The contour with the highest magnitude value corresponds to the invoice. By determining the coordinates of the four corners of this contour, we can crop the invoice image from the surrounding background.

Our cropping method involves adjusting the perspective of the invoice image (Fig. 4) to a frontal view, ensuring that the text is presented in a more straightforward and easily readable manner. While this technique may not offer the same level of accuracy as machine learning models specifically trained for this task, it provides a versatile solution that can be applied to diverse scenarios with similar requirements.

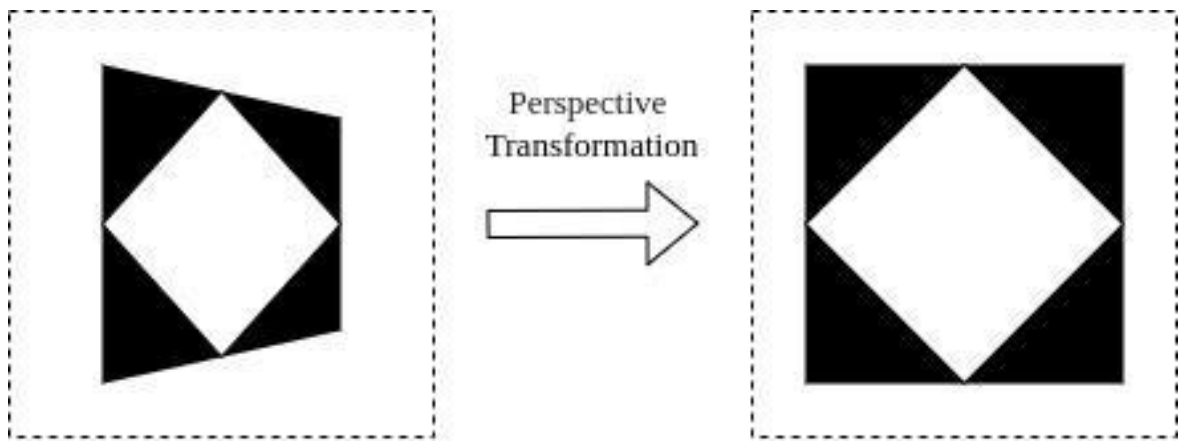

Fig. 4. Perspective transformation

#### 2. *Text detection*

The text detection problem shares similarities with object detection, but the focus is on detecting text instead of general objects. Deep learning architectures like Fast R-CNN, SSD, or YOLO can be applied to this problem, particularly their newer and more powerful versions. However, these models are typically more effective in detecting larger objects and may encounter challenges when dealing with small-sized objects, which is often the case with text data.

To address this issue, in this paper, we proposed a method called Segmentation-based, which treats text detection as object segmentation. The approach used is Real-time Scene Text Detection with Differentiable Binarization (DB) model [12].

Previous research models typically employ a fixed threshold to distinguish between background and object pixels in the probability map generated by the model. However, setting a fixed threshold introduces inflexibility and requires time-consuming efforts to determine the appropriate threshold value. To tackle this problem, the DB model allows learning the threshold value during training by fine-tuning the following equation [12]:

$$B_{i,j} = \frac{1}{1 + e^{-k(P_{i,j} - T_{i,j})}} \qquad (1)$$

Here, $B_{i,j}$ is the binary map, P is the probability map, T is the dynamic threshold map learned from the model, k is a gain index. The dynamic threshold map is derived from the feature map of the segmentation model. It is created using the same feature map as the probability map. However, instead of indicating the probability of a pixel belonging to the text object, it indicates whether the pixels belong to the surrounding edge of the object or not. This information assists the model in learning longer sentences with improved accuracy and more precise coverage.

By leveraging the pre-trained module and fine-tuning it with our specific dataset, we can take advantage of the knowledge and features learned from a larger corpus, while also tailoring the model to the specific requirements of our invoice dataset.

#### 3. *Text Recognition*

In this task, we use the Scene Text Recognition with a Single Visual Model (SVTR) method [13]. This method is effective in capturing semantic relationships between characters while optimizing training time and accuracy.

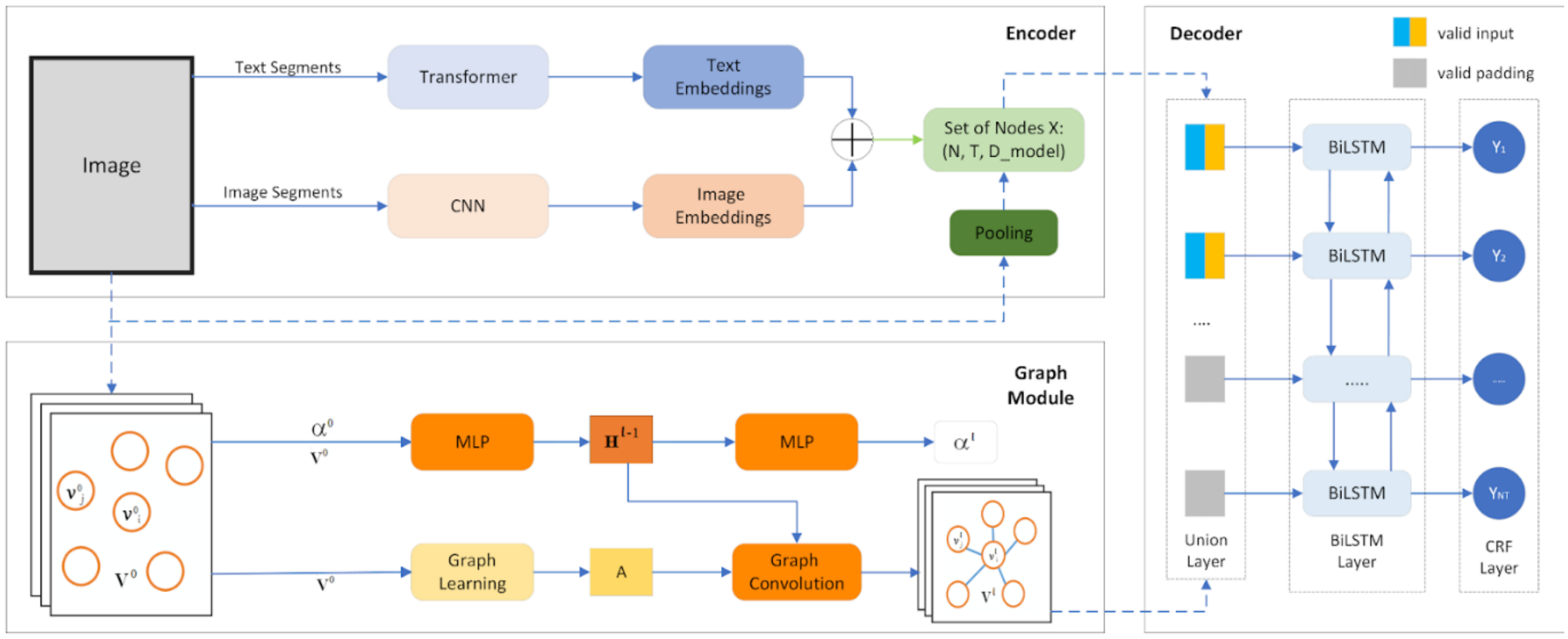

Fig 5. Overview of PICK

The SVTR architecture takes the positions of text regions as input and generates recognized text as output. The input consists of bounding box information obtained from a text detection model. The model architecture starts with an embedding layer called Patch Embedding, which divides the input image into a sequence [13].

The divided sequences then go through three layers, each composed of two main modules: Mix Block and Merge Block. A reliable word recognition model requires two crucial capabilities: effectively representing the correlation between details within the same character and accurately capturing the distinct characteristics of each character. Mix Block combines two types of blending: global mixing, which captures the relationship between non-text and text elements, and local mixing, which captures the relationship between strokes within the same character. These modules are designed to extract feature information at various scales, preventing information redundancy across layers. The information is then passed to Merge Block, which performs feature extraction at multiple scales and effectively eliminates information redundancy. Since the convolution layers used in Merge Block can lead to information loss and feature degradation when working with very small feature matrices [13], the model employs Combining Block at the end of network.

### 4. *Key Information Extraction*

The key information extraction method uses in this paper is Processing Key Information Extraction from Documents using Improved Graph Learning-Convolutional Networks (PICK) [14]. This method takes invoice images as input along with the detected text regions and the corresponding recognized text. This enables the model to learn both the structural features and the textual content present in each invoice.

PICK is considered a powerful and effective approach for extracting information with diverse structures from text [14]. It leverages three modules that work in synergy (Fig. 5): a graph learning module, which efficiently adjusts the model to capture the relationships between the extracted regions; an encoder; and a decoder, which extract features from both the text and the images to effectively represent their relationships. By encompassing features from a wide range of information, PICK achieves superior performance compared to other extraction models.

The use of PICK allows for comprehensive and accurate extraction of key information from documents, especially those with complex structures. The model has ability to capture relationships and represent features from both text and images contributes to its effectiveness in handling diverse document layouts and extracting the desired information accurately.

## IV. PERFORMAMCE EVALUATION

### A. *Test methodology*

The evaluation of the processed images is primarily conducted through visual observation since our image processing method relies solely on conventional image processing algorithms. This evaluation approach allows us to assess the credibility of the processed invoice images by identifying any possible misclassifications or errors. Fig. 6 provides a visual representation of the data after undergoing the preprocessing stage, showcasing the progress achieved in enhancing the quality and clarity of the invoice images.

Fig.6 shows that there are still instances where the background removal process is not entirely successful, and the alignment is not optimal. This issue, which involves accurately identifying invoices in object detection tasks, poses a considerable challenge. Even deep learning models encounter difficulties in detecting objects in images that are occluded or have poor lighting conditions, among other factors.

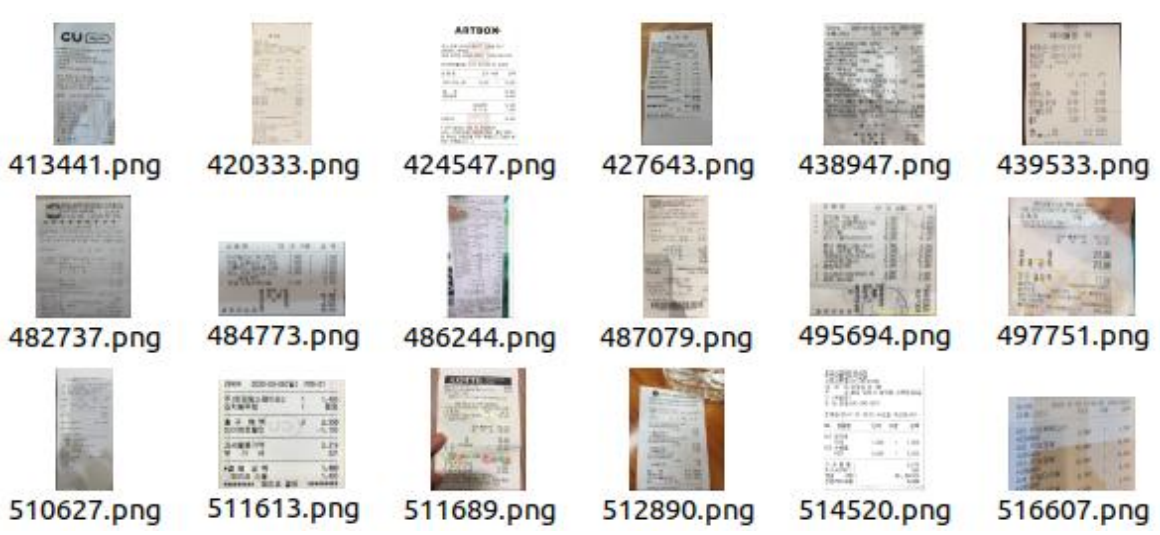

Fig. 6. Image obtained after processing money data

Nonetheless, considering the overall performance of our method, it provides reliable results by achieving a balance between speed and accuracy when implemented within the system. Despite the remaining challenges, our approach ensures a satisfactory level of performance, and the benefits of its speed and accuracy outweigh the limitations posed by certain image conditions and complexities. By employing our method, we can process a substantial amount of invoice data and extract the necessary information efficiently.

### B. *Text Detection assessment*

We will evaluate the system through IoU score [15] based on metrics, including Precision, Recall, and F1-Score. After fine-tuning with our dataset, which took approximately 151 consecutive hours on a CPU-based computer, the results were recorded in Table 1.

Table 1: IoU prediction measurement of the proposed model with pre-training of PPOCR

| IoU | PPOCR | | | Proposed method | | |
|---|---|---|---|---|---|---|
| | P | R | F1 | P | R | F1 |
| **0.5** | 0.78 | 0.9 | 0.84 | 0.84 | 0.91 | 0.87 |
| **0.75** | 0.62 | 0.72 | 0.67 | 0.7 | 0,76 | 0,73 |
| **0.8** | 0.53 | 0.61 | 0.57 | 0.62 | 0.69 | 0.64 |
| **Avg** | 0.64 | 0.74 | 0.69 | 0.72 | 0.79 | 0.75 |

*P: Precision, C: Recall, F1: F1-Scores*

Table 1 evidence that the fine-tuned model surpasses the accuracy of the pre-trained model obtained from the paddle ecosystem (PPOCR) [16]. Table 1 also shows that increasing the IoU threshold leads to a decrease in model accuracy. Specifically, when raising the threshold to 0.75, our method experiences a 14% decrease in accuracy, and the decrease is even more pronounced for the PPOCR model. When the threshold is set to 0.8, there is an approximate 10% decrease in accuracy.

This decline in accuracy can be led to several factors. One of the reasons is that the model was not trained for a sufficient duration due to limited data availability. Besides that, the limitations in equipment and time prevented us from further extending the training process. Additionally, accurately determining the positions of small text regions within invoices poses a challenge. Nevertheless, the overall results demonstrate a significant improvement compared to the initial pre-training, with an average accuracy increase of 6%.

In practical applications, the model exhibits stable and accurate detection of text regions. It can precisely capture text regions in most images, except for some challenging cases such as distorted, skewed, or poorly delineated images. However, this model is capable of accurately detecting text even in cases of blurry text or low-light images.

### C. *Text Recognization assessment*

In this step, training or fine-tuning is not involved, and instead, we utilize a pre-trained model from PPOCR. To assess the model's accuracy, we compiled a dataset of Korean text extracted from the AIHub website. The dataset comprises 6,922 images containing text segments with varying sizes and complexities.

The model achieved an accuracy score of 0.84, evaluated based on character accuracy. Overall, the model's performance is commendable. However, the real-world accuracy could potentially be higher, but it is influenced by the dataset preparation, which includes numerous special characters such as ■, 【, 「, and so on. These characters directly impact the model's accuracy. When examining the predictions for each Korean character, the accuracy is nearly perfect for high probability thresholds, with an average accuracy of approximately 0.9.

When applied to the current task, although the model performs well, there are still various limitations due to the presence of low-quality images, blurred or distorted characters, and images with deformations. Therefore, this model still faces several challenges in these scenarios.

### D. *Key Information Extraction assessment*

The information extraction process involves utilizing the labeled box positions, predicted transcriptions, and required information fields obtained from the previous steps. The information fields include details about purchased items, the total amount spent, and the time of purchase. To train the information extraction step, we use a combination of images, labeled box positions, and transcriptions as input data. These data are then trained on Google Colab using a GPU until the loss value no longer decreases. After continuous training for nearly 5 hours, the following results were obtained.

Table 2: Loss of graph-learning and CRF layer

| GL LOSS | CRF LOSS |
|---|---|
| 0.213 | 0.252 |

Table 3: Representation of information extraction results

| | mEP | mER | mEF |
|---|---|---|---|
| **Item** | 0.87 | 0.86 | 0.86 |
| **Total** | 0.85 | 0.71 | 0.77 |
| **Time** | 0.79 | 0.67 | 0.72 |
| **Avg** | 0.86 | 0.82 | 0.84 |

(mEP – mean entity precision, mER -mean entity recall, mEF- mean entity F1-score)

Table 2 shows that the loss values of the graph function and the CRF classifier have decreased significantly, indicating the model's ability to understand the invoice structure and classify the information. However, the results in Table 3 shows that the accuracy of the model varies across different fields. The recognition accuracy for the "item" field is relatively high, with an F1-Score of 86%. On the other hand, the accuracy for the other two fields is lower, at 77% and 72% respectively.

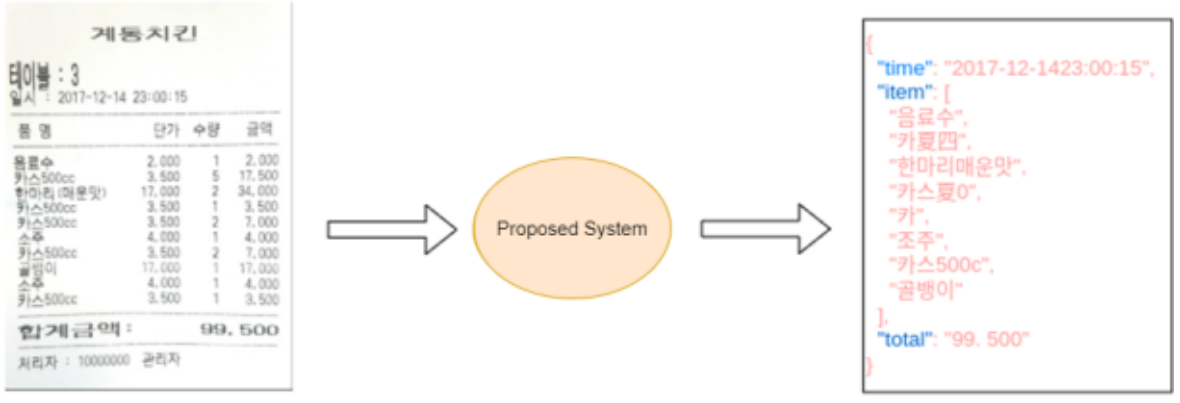

Figure 7: Final result on the system

There are several factors that affect performance uneven across fields. Firstly, the data used for training is primarily focused on purchased product information, leading to a limited representation of the other two fields. Secondly, the accuracy of the text recognition step also plays a role in the overall accuracy of the model. Inaccuracies in the recognized text can affect the learning ability of model and lead to reduced accuracy in information extraction.

Overall, the results demonstrate a satisfactory F1-Score of around 84%, ensuring the accuracy of the system. The model performs well when applied to invoices with good image quality and relies on the performance of previous models to provide accurate results.

### E. Overall system

The deployed system comprises multiple models, namely a text detection model, a text recognition model, and an invoice information extraction model (KIE), along with pre- and post-processing methods. These models have substantial sizes, with the information extraction model alone occupying 1.1GB of space. Consequently, initializing the program and loading the models may require a notable amount of time.

Fig. 7 provides a visual representation of the final output produced by the system. On average, the information extraction process takes approximately 5-10 seconds to complete, although the specific duration may vary depending on the length and complexity of the information contained within each invoice.

## V. Conclusion

In this paper, to achieve a practical and high accuracy model for extracting information from invoices with Korean language, we proposed a novel OCR model in which a completed dataset is collected and used to train in a complete system flow, input is Korean invoice image is firstly pre-proposed with a perspective transformation to remove the background, edit the view of the invoice. Next is the stages of detecting and recognizing the text appearing in the invoice. Finally, is to extract the important information contained in the invoice. Performance evaluation show that the proposed OCR model achieved higher accuracy results when compared with relevant PPOCR method, notably with about 87% in F1-Score while asking for negligible processing time. In the future, this system can completely improve and bring higher accuracy. We will proceed to collect more data, improve accuracy for each model. In addition, the system is completely suitable for practical applications, businesses or individuals can use for personal purposes.